# SSDOnt: an Ontology for representing Single-Subject Design Studies

I. Berges* (University of the Basque Country, UPV/EHU)

J. Bermúdez (University of the Basque Country, UPV/EHU)

A. Illarramendi (University of the Basque Country, UPV/EHU)

{idoia.berges, jesus.bermudez, a.illarramendi}@ehu.eus

* Corresponding author:

    Idoia Berges

    Computer Science Faculty, University of the Basque Country, UPV/EHU

    Paseo Manuel de Lardizabal, 1, 20018 Donostia – San Sebastián, Spain

    e-mail: idoia.berges@ehu.eus

**Summary**

**Background:** Single-Subject Design is used in several areas such as education and biomedicine. However, no suited formal vocabulary exists for annotating the detailed configuration and the results of this type of research studies with the appropriate granularity for looking for information about them. Therefore, the search for those study designs relies heavily on a syntactical search on the abstract, keywords or full text of the publications about the study, which entails some limitations.

**Objective:** To present SSDOnt, a specific purpose ontology for describing and annotating single-subject design studies, so that complex questions can be asked about them afterwards.

**Methods:** The ontology was developed following the NeOn methodology. Once the requirements of the ontology were defined, a formal model was described in a Description Logic and later implemented in the ontology language OWL 2 DL.

**Results:** We show how the ontology provides a reference model with a suitable terminology for the annotation and searching of single-subject design studies and their main components, such as the phases, the intervention types, the outcomes and the results. Some mappings with terms of related ontologies have been established. We show as proof-of-concept that classes in the ontology can be easily extended to annotate more precise information about specific interventions and outcomes such as those related to autism. Moreover, we provide examples of some types of queries that can be posed to the ontology.

**Conclusions:** SSDOnt has achieved the purpose of covering the descriptions of the domain of single-subject research studies.

1. Introduction

Single-Subject Design (SSD) has been widely used in educational [1] and social [2,3] settings during the past decades. Among other advantages, this approach provides a quick feedback about the effects of a treatment [2] and avoids missing each individual's experience [3], as it happens in group studies where the focus is put on group average. More recently, this paradigm has also been applied in other areas such as biomedicine [4] or physical theraphy [5]. In fact, more than 500 references of SSD studies published during the past 10 years can be found in PubMed [6].

It is well known that physicians often resort to electronic sources in order to search for specific studies in their field of interest. Nowadays these searches are usually performed by means of a syntactical search on the abstract, full text or keywords of the studies. In particular, in the PubMed and Cochrane [7] context, investigators often express their research questions using the PICO mnemonic [8] or CTSearch [9] with its interactive tag cloud display. Moreover, the PubMed Clinical Queries interface [10] allows to narrow the search for study designs. However, such faceted searches alone are often insufficient to support an appropriate granularity retrieval task. Those approaches pose difficulties, for example, for searching for SSD studies where the participants are within a specific age range, where the baseline phase comprises a specific number of sessions, or where the study has a specific structure. Fortunately, semantic technologies, such as ontologies, can play a relevant role in this scenario and help overcome these issues. Ontologies are knowledge representation artifacts that can be expressed by highly expressive (description) logic axioms capable of representing conceptual knowledge (i.e. classes of things) involving relevant properties of those things and relationships among them, in order to provide automatic reasoning. A significant corpus of ontologies has been developed for the medicine domain [11]. Among them, we can mention the Clinical Trials Ontology (CTO) [12], which provides a classification of clinical trial study types, without logical descriptions of these study types and their components, the Ontology for Biomedical Investigations (OBI) [13] that covers all phases of the biomedical investigation process, such as planning, execution and reporting, and finally, the Ontology of Clinical Research (OCRe) [14], which contains descriptions for the planning, execution and analysis of clinical research studies and trials. Neverthess in all of them the part relative to SSD studies lacks enough descriptors and appropriate granularity for looking for relevant information about them.

2. Objectives

The goal of this paper is to present SSDOnt, an OWL 2 DL [15] specific purpose ontology for describing and annotating SSD studies. The ontology covers the main features of the most popular types of SSD studies, as well as their main components, such as phases, intervention types, outcomes, results and relevant facts (e.g. age, condition) of the participants. SSDOnt tries to be a friendly artifact for those users interested in SSD studies. Those studies are classified in SSDOnt by their design characteristics. Thanks to these descriptions, complex queries about studies can be asked using DL Queries [16] or SPARQL [17]. Since the field of SSD studies is broad, the current version of SSDOnt provides a set of common classes and properties which can be extended for specific scenarios. As a proof of concept, we extend SSDOnt for the case of the autism spectrum disorder in children and youth.

3. **Methods**

The ontology was implemented using the NeOn Methodology framework [18]. Following the guidelines for the Ontology Requirements Specification Document (ORSD), scope, intended uses and competency questions were specified (see an excerpt of this document in Table 1, more details in [19]). The terms to describe SSD studies and their components that appeared in the conceptualization phase were selected from reference literature about the domain [1-5].

**Table 1 Excerpt of the ORSD for SSDOnt**

| Purpose | To provide a reference model for representing Single-Subject Design studies |
|---|---|
| Scope | The most typical SSD types found in the literature are considered. Moreover, as a proof of concept, the core ontology will be extended with information regarding practices for children, youth and young adults with Autism Spectrum Disorder (ASD). |
| Intended users | Physicians, researchers, or anyone interested in SSD. |
| Intended uses | Use 1: To annotate SSD studies and its components.<br>Use 2: To search for specific SSD studies. |
| Competency questions | - What is the type of the SSD study? (e.g across outcome multiple-baseline design)<br>- Which is the condition/pathology being studied? (e.g autism)<br>- Which intervention is used? (e.g peer-mediated intervention)<br>- What is the age of the participants?<br>- Retrieve SSD studies regarding [condition] (e.g regarding ASD)<br>- Retrieve SSD studies regarding [condition] in people younger/older than [age] //between [age1] and [age2] (e.g people between 1 and 3 years old) |

|  | - Retrieve SSD studies where [intervention] was used (e.g scripting) |
|  | - Retrieve multiple-baseline design studies where an across-{setting\|subject\|outcome} approach was taken. |

A Description Logic [20] was used to describe the classes and properties that are needed to answer the competency questions listed in the ORSD (see section Results). The formal model was then implemented in the ontology language OWL 2 DL using Protégé 5.0 [21]. SSDOnt imports the Bibliographic Ontology, BIBO [22], a reference ontology for annotating bibliographic resources, to link each SSD study with any publications that have derived from it, and several mappings have been defined between terms in SSDOnt and some well-known ontologies such as the Semanticscience Integrated Ontology (SIO) [23], Open Biomedical Ontologies (OBO) [24] and OCRe. In [19] those mappings can be consulted.

**Ethical guidelines**

All the data concerning a particular patient that are presented in this research are synthetic.

## 4. Results

The core of the SSDOnt ontology contains 44 classes and 33 properties (excluding those of BIBO). This relatively small number of terms make the ontology a friendly artifact for users interested in SSD studies. Next, we introduce first some of its main classes and properties for the semantic annotation of SSD studies and its main components. Then we show how the competency questions defined in the ORSD can be answered using the ontology.

*4.1 Annotation of SSD studies*

The first intended use of the ontology is to serve as a reference vocabulary for annotating SSD studies. The main class *SingleSubjectDesign* represents all the SSD studies, and has been divided into four subtypes of studies: *SimpleDesign*, *WithdrawalDesign*, *MultipleBaselineDesign* and *AlternatingTreatmentDesign*. These classes have been further expanded to represent more specific concepts, such as ABAB designs (*ABAB_Design*), which is a subclass of *WithdrawalDesign*, or *AcrossSettingMBD*, which is a subclass of *MultipleBaselineDesign* (see Figure 1a).

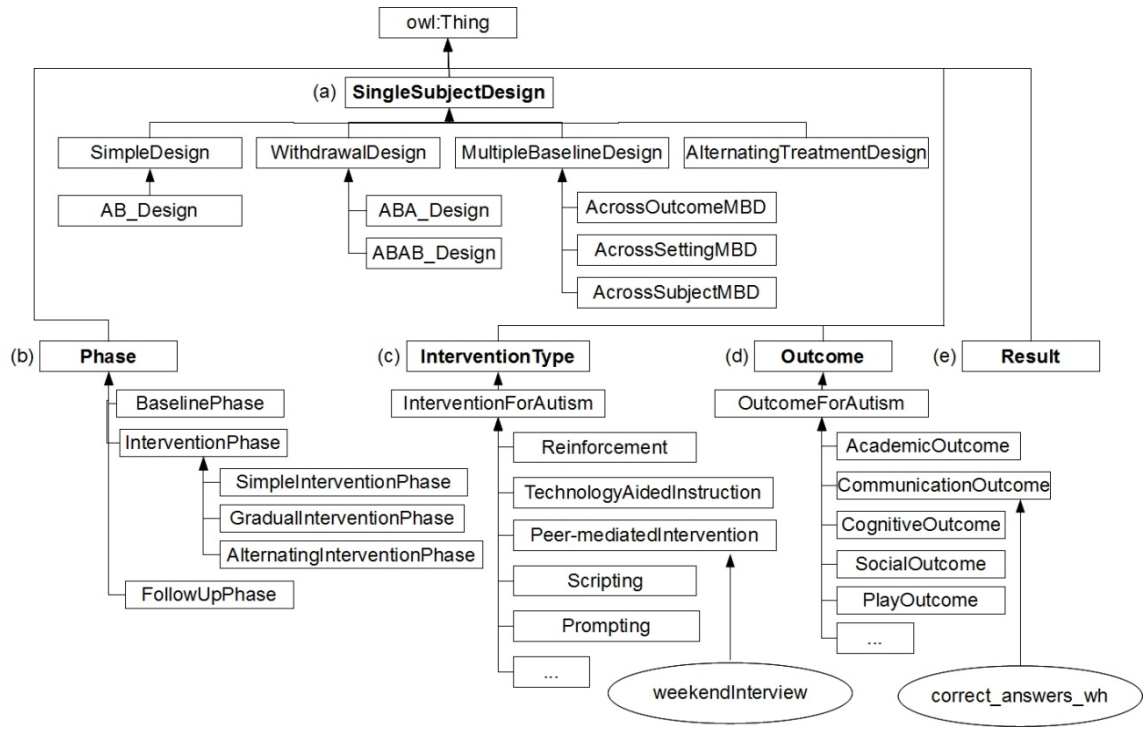

**Figure 1 Excerpt of the SSDOnt ontology with specific subclasses for autism**

SSD studies are developed in phases (see Figure 1b). Thus, three types of phases have been defined: *BaselinePhase* (for representing phases where baseline measurements are taken without intervention), *InterventionPhase* (phases where intervention is applied) and *FollowUpPhase* (optional phase for post-treatment follow-up).

For example (Figure 2a), an ABAB design is defined as a withdrawal design which either has four phases (baseline, intervention, baseline and intervention phases) or five (baseline, intervention, baseline, intervention and follow up). It is worth noting that the model can be easily extended to cover other withdrawal designs such as *ABABABAB_Design*, just by defining new subclasses of *WithdrawalDesign*.

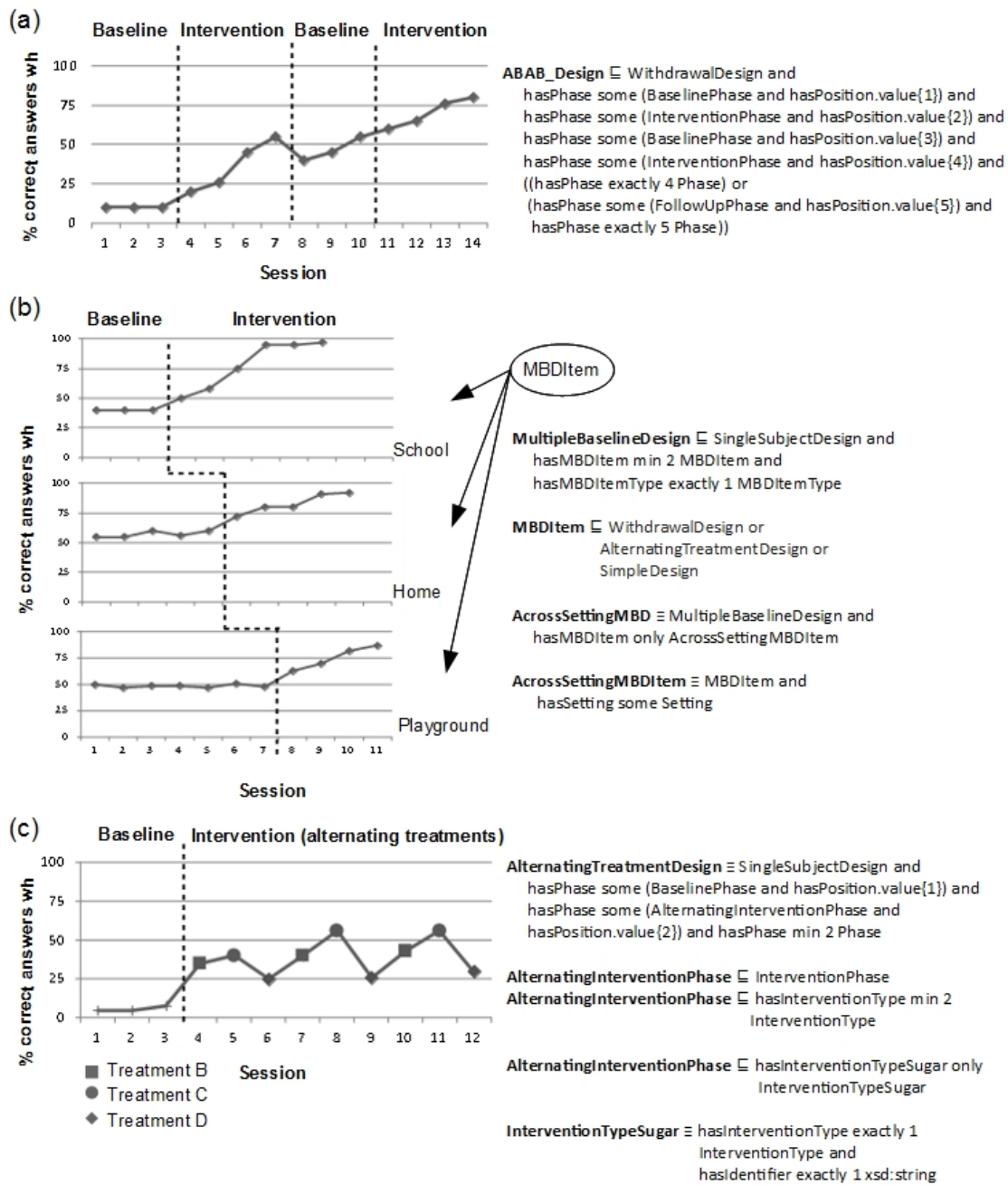

**Figure 2** Examples of types of SSD studies and their formal descriptions: (a) ABAB design, (B) Across-setting multiple-baseline design, (C) Alternating treatment design.

One more complex definition is that of multiple baseline designs. A multiple baseline design is a single-subject design which has at least two substudies (*MBDItem*) carried out in parallel (see Figure 2b). The *MBDItem*s can follow the structure of either a *WithdrawalDesign,* an *AlternatingTreatmentDesign,* or a *SimpleDesign*, (all MBDItems within the same study must follow the same structure. This is specified with the *hasMBDItemType* property). Apart from the length of the baseline phase, they differ from each other in one of the following dimensions: the outcome (i.e what is being measured; e.g the % of correct answers to Wh-questions, the number of tantrums in one day), the setting (i.e where it is being studied; e.g. at school, at home) or the subject (i.e who is being studied; e.g. Paul, Mary). Thus, the class *MultipleBaselineDesign* has been specialized to accommodate three subclasses of multiple baseline designs: *AcrossOutcomeMBD*, *AcrossSettingMBD* and *AcrossSubjectMBD*. For example, in the *AcrossSettingMBD* in Figure 2b, the same subject is examined for the same target outcome (% of correct answers to Wh-questions) but in different settings (home, school, playground).

Finally, an alternating treatment design (Figure 2c) begins with a baseline phase, which is followed by an intervention phase where at least two different treatments (interventions) are applied. In other words, two or more treatments are alternated during the intervention phase, unlike in the other types of designs, where the same intervention is applied during the whole intervention phase. Thus, a subclass of *InterventionPhase* has been created, namely *AlternatingInterventionPhase* to represent this particularity (see Figure 1b).

*4.2 Annotation of types of intervention and outcomes*

Other relevant classes in the ontology are *InterventionType* and *Outcome*. Class *InterventionType* (Figure 1c) represents the different types of intervention that can be used within SSD studies (i.e. the independent variable). Since the nature of the intervention depends on the target condition for which is applied, no subclasses of it have been defined in the core ontology. Even so, instances of *InterventionType* can be created directly, but when using it in an specific context, it would be advisable to create a classification of typical intervention types related to that target condition. As proof of concept, we provide a classification of typical intervention types related to autism, extracted from [25]. Specific intervention actions will be instances of these intervention type classes. For example, *weekendInterview* is an instance of *Peer-mediatedIntervention*, in which children gather in small groups and ask each other questions about their weekend plans.

Class *Outcome* (Figure 1d) represents the variable of interest that is being measured in the study (i.e. the dependent variable). With property *inFormOf*, we can indicate whether the results measured for that outcome are represented as a *percentage*, a *magnitude*, *duration*, a *frequency* or an *interval* between events. Moreover, the outcome is once again tightly linked to the nature of the study, so no subclasses of Outcome are provided in SSDOnt. However, as in the previous case, we show a classification of types of outcomes related to autism [25], which will be populated with instances such as *correct_answers_wh,* which is an instance of *CommunicationOutcome* to represent the % of correct answers to Wh-questions.
,
*4.3 Annotation of results*

Each of the measurements taken during the study will be an instance of class *Result*. Each result is characterized by its value (*hasValue*), the *Instant* in which it was measured (*occursIn*), the *InterventionType* that was being used at the moment (*hasInterventionType*), if any, and the *Phase* to which it belongs (*isResultOfPhase*). For example, let us assume that Paul (fictional patient) is a 7 year - 4 month old boy diagnosed with autism at age 3. He has participated in a SSD study to detect whether interviews with his peers about their weekend plans can improve his ability to answer Wh-questions. The results of the study are those of Figure 2a. Some of the annotations that are created can be seen in Figure 3 (in Turtle [26] syntax[1]).

```
ssd:paul a ssd:Participant ;
    ssd:hasCondition ssd:autism ;
    ssd:hasGender ssd:male ;
    ssd:hasAge _:age01 ;
    ssd:diagnosedAtAge _:age02 .

_:age01 a ssd:AgeDescription ;
    ssd:years 7 ;
    ssd:months 4 .

_:age02 a ssd:AgeDescription ;
    ssd:years 3 .

ssd:ssd01 a ssd:ABAB_Design ;
    ssd:hasParticipant ssd:paul ;
    ssd:hasOutcome aut:correct_answers_wh ;
    ssd:hasPhase ssd:ph01 ;
    ssd:hasPhase ssd:ph02 ;
    .... .

aut:correct_answers_wh a aut:CommunicationOutcome .
ssd:ph01 a ssd:BaselinePhase ;
    ssd:hasPosition 1 .

ssd:ph02 a ssd:SimpleInterventionPhase ;
    ssd:hasPosition 2 ;
    ssd:hasInterventionType aut:weekendInterview .

aut:weekendInterview a aut:Peer-mediatedIntervention

ssd:res01 a ssd:Result ;
    ssd:hasValue 10.1 ;
    ssd:occursIn _:inst01 ;
    ssd:isResultOfPhase ssd:ph01 .

_:inst01 a ssd:Instant ;
    ssd:hasValue 1 .

ssd:res02 a ssd:Result ;
    ssd:hasValue 10.1 ;
    ssd:occursIn _:inst02 ;
    ssd:isResultOfPhase ssd:ph01 .

_:inst02 a ssd:Instant ;
    ssd:hasValue 2 .
....

ssd:res04 a ssd:Result ;
    ssd:hasValue 20.4 ;
    ssd:occursIn _:inst04 ;
    ssd:isResultOfPhase ssd:ph02 ;
    ssd:hasInterentionType aut:weekendInterview .

_:inst04 a ssd:Instant ;
    ssd:hasValue 4 .
...
```

**Figure 3 Some annotations for the study in Figure 2a**

### 4.4 Querying for SSD studies

Once the studies have been annotated, querying for specific studies or for information within those studies can be performed by means of DL Queries or SPARQL queries. Thanks to the richness of annotations allowed by SSDOnt, both simple and complex queries can be posed. Next we show some examples:

---

[1] In SSDOnt, unique identifiers have been used to create the URIs. However, for the sake of readability, we show here the labels of the corresponding terms.

***DL Query (simple):*** *Retrieve the results of phase ph01:*

*Result* and *isResultOfPhase* some {*ph01*}

***DL Query (complex):*** *Retrieve across-setting multiple-baseline studies where the participant is younger than 10 years old and where at least one of the observed settings is 'school':*

*AcrossSettingMBD* and *hasParticipant* some (*Participant* and *hasAge* some (*years* some xsd:int[<10])) and

*hasMBDItem* some (*AcrossSettingMBDItem* and *hasSetting* value *school*)

***SPARQL:*** *Retrieve the best result obtained in the intervention phase of AB studies for improving answering to Wh-questions, where any form of peer-mediated intervention is used:*

```
PREFIX ssd: <http://bdi.si.ehu.es/bdi/ontologies/SSDOnt/SSDOnt#>
PREFIX aut: <http://bdi.si.ehu.es/bdi/ontologies/SSDOnt/SSDOntAutism#>
SELECT ?study ?interType ?val
WHERE {
?study a ssd:AB_Design ; ssd:hasOutcome aut:correct_answers_wh ; ssd:hasPhase ?ph .
?ph a ssd:SimpleInterventionPhase ; ssd:hasInterventionType ?interType .
?interType a aut:Peer-mediatedIntervention .
?res ssd:isResultOfPhase ?ph ; ssd:hasValue ?val
} order by DESC(?val) LIMIT 1
```

A user-friendly interface will allow non-expert users to query about studies annotated with SSDOnt.

## 5. Discussion

In this paper we have introduced an ontology that captures information concerning SSD studies, which can yield numerous benefits for investigators interested in expressing scientific research questions about them. Due to the use of a logic-oriented ontological approach, our proposal can support computational reasoning and thus it allows to go a step further than other proposals for clinical research (e.g. CDISC [27]), more oriented to achieve interoperability in data exchange and application development. In the case of the ClinicalTrials.gov [28] model, although it is relevant because it is the world's largest collection of study design information, its modeling of studies design lacks depth [29]. Furthermore, while there also machine learning approaches exist that try to determine the relevance of studies to a search query (e.g. [30]) many of them require training a classifier against a hard-coded gold standard, which limits their scalability.

In order to build SSDOnt we have followed the well-known NeOn methodology. Moreover, the terms to describe SSD studies and their components that appear within the ontology have been selected from reference literature about the domain [1-5], in order to be appropriate for researches in the field.

Annotating studies using an ontology such as SSDOnt enhances the retrieval of information, since it broadens the spectrum of queries that can be answered. Structure features of studies can be queried with adecuate granularity. Moreover, the ontology provides a foundation to help in the design of Single-Subject research studies.

The ontology was assessed against several quality criteria for ontology evaluation described in [31]: *accuracy and conciseness* (the definitions in SSDOnt were created after a thorough research about single-subject research, so they conform to the expert's knowledge about that domain and do not contain irrelevant terms), *adaptability* (SSDOnt can be easily extended to cover non-usual designs and any condition), *clarity* (all terms in SSDOnt include the standard annotations *rdfs:label* and *rdfs:comment* to indicate their meaning), *completeness* (SSDOnt can answer all the competency questions specified in the ORSD), *consistency* (no inconsistencies were found when performing reasoning on the ontology)., and *computational efficiency* (reasoning can be performed over SSDOnt in negligible time using a reasoner such as Fact++ [32]) Concerning this last criteria a test was carried out creating annotations for 1000 bogus SSD studies, with 186,679 axioms and 51,508 individuals and were performed evaluation of DL Queries and SPARQL queries. For DL Queries we used Protégé. It took the reasoner Fact++ 38 seconds to classify the ontology the first time, and few seconds to answer DL Queries such as those in section 4.4. As for SPARQL queries, we loaded the aforementioned annotations into the GraphDB [32] triple store, and queries such as the one in section 4.4 took less than a second to be processed. The file containing these annotations can be found in [19].

6.  **Conclusions**

Single-subject research has proven its usefulness in several domains such as biomedicine. This usefulness can be enhanced by providing a computationally-tractable specification of the configurations of SSD studies and their results, with a suitable terminology and granularity for annotation and searching. In this paper we have presented SSDOnt, an OWL 2 DL specific purpose ontology for annotating SSD studies which allows to express complex queries and retrieve information about the components of those studies. The ontology complies with the most usual quality criteria considered when performing ontology evaluation. The source file containing the ontology, as well as its documentation and some annotated examples, can be found in [19].

**7. References**

1. Kennedy CH. Single-case Designs for Educational Research. 1st ed:New York: Pearson; 2004.
2. Lammers WJ, Badia P. Fundamentals of Behavioral Research. Belmont, CA: Wadsworth Inc.; 2005.
3. Engel RJ, Schutt RK. The Practice of Research in Social Work. 4th ed:Thousand Oaks, CA: SAGE Publications, Inc.; 2017.
4. Janosky JE, Leininger SL, Hoerger MP. Single Subject Designs in Biomedicine. Dordrecht Heidelberg London New York: Springer; 2009.
5. Carter R, Lubinsky J. Rehabilitation Research: Principles and Applications. 5th ed:St. Louis, MO: Elsevier; 2015.
6. PubMed. https://www.ncbi.nlm.nih.gov/pubmed. Accessed 20 September 2017.
7. Cochrane. http://www.cochranelibrary.com/. Accessed 20 September 2017.
8. Richardson WS, Wilson MC, Nishikawa J, Hayward RS. The well-built clinical question: a key to evidence-based decisions. ACP J Club. 1995;123:A12–3
9. Hernandez M, Falconer SM, Storey M, Carini S, Sim I. Synchronized tag clouds for exploring semi-structured clinical trial data. Proceedings of the 2008 conference of the center for advanced studies on collaborative research: meeting of minds, Ontario, Canada, pp. 42–56 (2008)
10. PubMed Clinical Queries. https://www.ncbi.nlm.nih.gov/pubmed/clinical. Accessed 20 October 2017.
11. BioPortal. http://bioportal.bioontology.org/. Accessed 20 September 2017.
12. Clinical Trials Ontology, CTO. http://bioportal.bioontology.org/ontologies/CTO. Accessed 20 September 2017.
13. Ontology for Biomedical Investigations, OBI. http://obi-ontology.org/ Accessed 20 September 2017
14. Ontology of Clinical Research, OCRe. https://bioportal.bioontology.org/ontologies/OCRE. Accessed 20 September 2017.
15. OWL 2 DL. https://www.w3.org/TR/owl2-overview/. Accessed 20 September 2017.
16. DL Query. http://protegewiki.stanford.edu/wiki/DL_Query. Accessed 20 September 2017.

17. Harris S, Seaborne A, editors. SPARQL 1.1 Query Language. W3C Recommendation, 2013. https://www.w3.org/TR/sparql11-query/. Accessed 20 September 2017.

18. Suárez-Figueroa MC, Gómez-Pérez A, Fernández-López M. The NeOn Methodology for Ontology Engineering. In: Suárez-Figueroa MC, Gómez-Pérez A, Motta E, Gangemi A, editors. Ontology Engineering in a Networked World., pp. 9-34. Berlin Heidelberg:Springer; 2012.

19. SSDOnt. Documentation. http://bdi.si.ehu.es/bdi/ontologies/SSDOnt/docs/SSDOnt.html. Accessed 20 September 2017.

20. Baader F, Calvanese D, McGuinness DL, Nardi D, Patel-Schneider PF, editors. The Description Logic Handbook: Theory, Implementation, and Applications. 2nd ed: Cambridge: Cambridge University Press; 2007.

21. Protégé. http://protege.stanford.edu. Accessed 20 September 2017.

22. D'Arcus B, Giasson F. Giasson F, editors. Bibliographic Ontology Specification, 2009. http://http://bibliontology.com/. Accessed 20 September 2017.

23. Semanticscience Integrated Ontology, SIO. https://bioportal.bioontology.org/ ontologies/SIO. Accessed 20 September 2017.

24. Open Biomedical Ontologies, OBO. www.obofoundry.org/. Accessed 20 September 2017.

25. Wong C, Odom SL, Hume K, Cox AW, Fettig A, Kucharczyk S, et al. Evidence-Based Practices for Children, Youth, and Yound Adults with Autism Spectrum Disorder. Chapel Hill: The University of North Carolina, Frank Porter Graham Child Development Institute, Autism Evidence-Based Practice Review Group; 2013.

26. Beckett D, Berners-Lee T, Prud'hommeaux E, Carothers G. Prud'hommeaux E, Carothers G editors. RDF 1.1 Turtle: Terse RDF Triple Language. W3C Recommendation, 2014. https://www.w3.org/TR/turtle/. Accessed 20 September 2017.

27. CDISC Protocol Representation Model. https://www.cdisc.org/standards/foundational/protocol. Accessed 20 September 2017.

28. ClinicalTrials.gov. A service of the U.S. National Institutes of Health. https://clinicaltrials.gov. Accessed 20 September 2017.

29. Roumiantseva D, Carini S, Sim I, Wagner TH. Sponsorship and Design Characteristics of Trials Registered in ClinicalTrials.gov. Contemporary Clinical Trials, 2013; 34:348–355

30. Wallace BC, Trikalinos TA, Lau J, Brodley C, Schmid CH. Semi-automated Screening of Biomedical Citations for Systematic Reviews. BMC Bioinformatics, 2010; 11:55

31. Vrandecic D. Ontology evaluation. Staab S, Studer R. editors. Handbook on Ontologies. International Handbooks on Information Systems, pp. 293-313. Berlin Heidelberg: Springer; 2009.

32. Tsarkov D, Horrocks I. Fact++ description logic reasoner: System description. Furbach U, Shankar N, editors. Automated Reasoning, Proceedings of the Third International Joint Conference, IJCAR 2006, Seattle, WA, USA. Lecture Notes in Computer Science, vol. 4130, pp. 292-297. Springer, Berlin Heidelberg (2006)
33. GraphDB. http://graphdb.ontotext.com/. Accessed 20 September 2017.